\RequirePackage{fix-cm}

\documentclass[twocolumn]{svjour3}          

\smartqed

\usepackage{subfig}
\usepackage{graphicx}
\usepackage{amsmath}
\usepackage{mathtools}
\usepackage{url}
\usepackage[linesnumbered,ruled]{algorithm2e}
\usepackage[sort&compress,numbers]{natbib}
\usepackage[colorlinks,allcolors=blue]{hyperref}
\usepackage{multirow}
\usepackage{subfiles}

\newcommand{\bigO}{\mathcal{O}}
\newcommand{\realR}{{\rm I\!R}}

\usepackage{pgfplots}
\usepackage{pgfplotstable}
\pgfplotsset{compat=1.11,
        /pgfplots/ybar legend/.style={
        /pgfplots/legend image code/.code={
        \draw[##1,/tikz/.cd,bar width=3pt,yshift=-0.2em,bar shift=0pt]
                plot coordinates {(0cm,0.8em)};},
},
}

\journalname{International Journal of Machine Learning and Cybernetics}

\def\makeheadbox{\relax}

\begin{document}

\title{DISCERN: Diversity-based Selection of Centroids for k-Estimation and Rapid Non-stochastic Clustering}

\author{Ali Hassani \and Amir Iranmanesh \and Mahdi Eftekhari \and Abbas Salemi}

\institute{
	A. Hassani \at
	Department of Computer Science, Shahid Bahonar University of Kerman, 76169-14111, Pajoohesh Square, Kerman, I.R. Iran. \\
	\email{alihassanijr1998@gmail.com}
	\and
	A. Iranmanesh \at
	Department of Computer Science, Shahid Bahonar University of Kerman, 76169-14111, Pajoohesh Square, Kerman, I.R. Iran. \\
	\email{amir.ir1374@gmail.com}
	\and
	M. Eftekhari \at
	Department of Computer Engineering, Shahid Bahonar University of Kerman, 76169-14111, Pajoohesh Square, Kerman, I.R. Iran. \\
	\email{m.eftekhari@uk.ac.ir}
	\and
	A. Salemi (Corresponding author) \at
	Department of Applied Mathematics and Mahani Mathematical Research Center, Shahid Bahonar University of Kerman, 76169-14111, Pajoohesh Square, Kerman, I.R. Iran. \\
	\email{salemi@uk.ac.ir}
	\\
}

\date{September 2020}

\maketitle

\begin{abstract}
One of the applications of center-based clustering algorithms such as K-Means is partitioning data points into $K$ clusters. In some examples, the feature space relates to the underlying problem we are trying to solve, and sometimes we can obtain a suitable feature space. Nevertheless, while K-Means is one of the most efficient offline clustering algorithms, it is not equipped to estimate the number of clusters, which is useful in some practical cases. Other practical methods which do are simply too complex, as they require at least one run of K-Means for each possible $K$. In order to address this issue, we propose a K-Means initialization similar to K-Means++, which would be able to estimate $K$ based on the feature space while finding suitable initial centroids for K-Means in a deterministic manner. Then we compare the proposed method, DISCERN, with a few of the most practical $K$ estimation methods, while also comparing clustering results of K-Means when initialized randomly, using K-Means++ and using DISCERN. The results show improvement in both the estimation and final clustering performance.
\end{abstract}

\section{Introduction}
\label{sec:introduction}
Due to the vast growth in data and data generation, machine learning methods have started to grow more rapidly than ever, in order to be able to catch up with this growth in data. Processing all of this information is usually divided into four categories: supervised learning, unsupervised learning, semi-supervised learning and reinforcement learning.
In supervised learning, the objective is known and the relationship between the input and the objective is the target of the learning process. In unsupervised and semi-supervised learning however, the objective is usually unknown. For instance, in supervised learning, emails can be classified into labels, examples of which are available, but in unsupervised learning, emails can be grouped into a certain number of groups based on their own properties. In semi-supervised learning, specific details are added to the unsupervised learning process or supervised cases are then applied to unsupervised measures.
One of the most prominent algorithms in unsupervised and even semi-supervised methods of machine learning is data clustering. It is a process of partitioning data into clusters, based on their similarity to each other. Many different clustering approaches for unsupervised learning and constrained clustering approaches for semi-supervised learning have been introduced and are actively being applied to practical tasks.

While clustering algorithms have many other usages as well, dividing data points into different clusters can only be as good as the feature space of the data points. Methods such as spectral embedding, kernel methods and representation learning methods such as deep learning have been employed in order to improve clustering algorithms by providing a better representation.
Deep learning methods have gained a great deal of attention recently, due to their strength in truly analyzing and later representing data. Deep learning has been applied to a great number of tasks, even unsupervised and semi-supervised learning tasks such as clustering and constrained clustering \cite{caron2018deep,zhang2019deep} as well as online clustering methods which aide representation learning \cite{gansbeke2020learning,caron2020unsupervised}.

In most offline partitioning clustering methods, the number of clusters is usually the initial parameter. This number can be unknown in practical cases, leaving the determination or estimation of this number open to discussion, and many researchers have proposed measures to address this problem. An instance of such a problem can be person re-identification and face grouping, where a set of images of people’s faces is given, and the objective is grouping by identity. In this problem, the number of people, which is the number of the groups, or in this case, clusters, is unknown.
K-Means is one of the most efficient offline clustering algorithms, and one the most widely used partitioning methods. These algorithms represent each cluster with center points or centroids and any given data point is then assigned to the nearest centroid’s corresponding cluster. K-Means has been widely used in subjects such as patent detection \cite{kim2012graph} and music recognition \cite{fang2015visual}, while being one of the most commonly used clustering methods. The original algorithm starts with $K$ random points in the data space as centroids. This is known as one of its primary downsides, as it makes the algorithm sensitive to this step, which is usually referred to as the \textbf{initialization phase}.
K-Means++ \cite{arthur2007k} was later introduced in order to initialize K-Means more effectively, but it is still dependant on the number of clusters. With the added complexity of K-Means++, while the algorithm is still efficient, running the algorithm multiple times along with K-Means itself in order to estimate the number of clusters is very inefficient. Moreover, the stochastic nature of the algorithms may lead to slightly different results each time, which would also affect this process.

In short, one of the issues of both  K-Means and K-Means++ is inefficiency when attempting to estimate the number of clusters through sequential runs with different parameters, especially when there is no known upper bound for that number. Given a set consisting of $n$ data points, the number of clusters can vary from $2$ to even as far as $n$. Another problem when using these methods is stability in results, as different runs with the same parameters can yield different results. While this can be seen an advantage in some cases, it can however be an issue when using these algorithms to divide the data points into clusters, similar to classification.
In order to mitigate these issues, we propose a new method which performs similarly to K-Means++, but does not require the number of clusters to be determined. This method can estimate the number of clusters while estimating the initial centroids which are then passed to K-Means. Moreover, this method is deterministic, yielding the same results regardless of the random state and the order of the data points.
We compare the performance and complexity of DISCERN to the most practically used methods for cluster estimation, and also compare the clustering performance in some supervised learning problems to K-Means and K-Means++.

The remainder of this paper is organized as follows: In Sect. \ref{sec:notation} the notation used in the paper is provided, Sect. \ref{sec:relatedwork} covers related work and section \ref{sec:method} presents the proposed approach. Section \ref{sec:experiments} covers the details of the experiments conducted on the proposed method, the results, and discussion on the results. We conclude and present possible future explorations in Sect. \ref{sec:conclusion}.

\section{Notation}
\label{sec:notation}
The notations used in this paper are as follows:
When clustering data into $K$ clusters, where each cluster is a set of data points and has a center point or centroid which is denoted as $z_l$:
\begin{equation}
    \label{eq:centroid}
    z_l = \frac{1}{|C_l|} \sum\limits_{x \in C_l}{x}.
\end{equation}
The initial centroids or $z_l$'s were originally selected randomly among data points. This process will be explained further in the next section.
Given the set of data points in the form of a matrix $X \in \realR^{n \times d}$ where $n$ is the number of data points, each data point $x_i$ can be labeled using a distance metric and the centroids of the $K$ clusters.
Equation \ref{eq:clusterassignment} assigns data points to clusters based on Euclidean distance and is used most commonly.
\begin{equation}
    \label{eq:clusterassignment}
    l_i = \operatorname*{argmin}_{1 \leq j \leq K} {\| x_i - z_j \|_2 }.
\end{equation}
Another form of clustering using cosine similarity, or spherical clustering is also used depending on the data space. Spherical cluster assignment is presented in \ref{eq:sphericalclusterassignment}.
\begin{equation}
    \label{eq:sphericalclusterassignment}
    l_i = \operatorname*{argmax}_{1 \leq j \leq K} {cos(\theta(x_i,z_j))}.\\
\end{equation}
Cosine similarity in \ref{eq:sphericalclusterassignment} can be expressed as:
\begin{equation}
    \label{eq:cosinedistance}
    cos(\theta(x,y)) = \frac{x^T y}{\| x \|_2 \| y \|_2} .
\end{equation}
Spherical clustering has been applied to text mining and document clustering \cite{gulnashin2019new,jain2018clustering}. The distance metric is therefore an important part of the definition of cluster assignment.

We denote $\circ$ as the element-wise (Hadamard) product of two matrices $H$ and $J$ of the same size, which is defined as:
\begin{equation}
\label{eq:hadamard}
H \circ J :=  [H_{ij} J_{ij}], \quad where \ H = [H_{ij}] , J = [J_{ij}]  .
\end{equation}

\section{Related Work}
\label{sec:relatedwork}
Since the proposed method in this paper is an initialization method for K-Means and is related to methods such as K-Means++, while also being related to the elbow and silhouette methods and X-Means for K-estimation, we present brief reviews on these methods in this section.

\subsection{K-Means}
As previously discussed, K-Means, and its variants are among the most widely employed clustering algorithms for their efficiency and performance. The original K-Means algorithm is presented in Algorithm \ref{alg:kmeans}.
\begin{algorithm}
	\SetKwInOut{Input}{Input}
	\SetKwInOut{Output}{Output}

	\Input{Dataset $X \in \realR^{n \times d}$, maximum number of iterations $t$, and the number of clusters $K$}
	\Output{Cluster assignments $L$}
	Initialize $Z \in \realR^{K \times d}$ with random values, $L \in \realR^{n \times 1}$ with zeros\;
	\For{$i = 1$ \KwTo $T$}
	{
		\For{$j = 1$ \KwTo $n$}
		{
			$l_j = \operatorname*{argmin}_{1 \leq k \leq K} {\| x_j - z_k \|_2}$
		}
		\For{$l = 1$ \KwTo $K$}
		{
			$n_l = | C_l |$\;
			$z_l = \frac{1}{n_l} \sum\limits_{x \in C_l}{x}$
		}
	}
	\caption{K-Means}
	\label{alg:kmeans}
\end{algorithm}
The computational complexity of this algorithm is $\bigO(ndKt)$ \cite{hartigan1979algorithm}, which when considering $K$ and $t$ constants is lower than the complexity of most other clustering algorithms. Some have also recently proposed methods to further improve the performance of K-Means \cite{sarma2013hybrid}, and others based their methods on this algorithm \cite{wang2019three,chen2018ordered}.
\subsection{K-Means++}
K-Means++ \cite{arthur2007k} aims to boost its performance by initializing the centroids more intuitively, and is presented in Algorithm \ref{alg:kmeanspp}.
\begin{algorithm}[!ht]
	\SetKwInOut{Input}{Input}
	\SetKwInOut{Output}{Output}

	\Input{Dataset $X \in \realR^{n \times d}$, maximum number of iterations $t$, and the number of clusters $K$}
	\Output{Cluster assignments $l \in \realR^{n \times 1}$}
	Initialize $Z \in \realR^{K \times d}$ with zeros\;
	$z_1 = $ select a random row of $X$\;
	\For{$i = 2$ \KwTo $K$}
	{
		Initialize $d \in \realR^{n \times 1}$ with zeros\;
		\For{$j = 1$ \KwTo $n$}
		{
			$d_j = \min\limits_{1 \leq k \leq i - 1}{\| x_j - z_k \|_2^2}$
		}
		$z_i = $ select a row of $X$ based on probability proportional to $d$
	}
	$L = \text{K-Means}(X,K,Z,t)$\;
	\caption{K-Means++}
	\label{alg:kmeanspp}
\end{algorithm}
Despite being over a decade old, this method is still considered one of the most prominent partitional clustering methods, as it has been applied recently in many cases such as recommendation systems \cite{cai_zhou_li_2019} and image processing \cite{solak2018new}.
This extension to K-Means also relies on the number of clusters given as input, on top of being stochastic. In practical cases, it may be suitable to attempt multiple runs of this algorithm, and select the best one as the optimum clustering.
The computational complexity of this algorithm without considering the K-Means complexity which comes afterwards is $\bigO(ndK^2)$.

\subsection{Spectral Clustering}
\begin{figure}
	\begin{tikzpicture}
		\begin{axis}[
    		width=0.485\textwidth,
    		height=6.5cm,
            scatter/classes={
                a={mark=o,draw=blue}, b={mark=o,draw=red}}]
            \addplot[scatter,only marks,
                scatter src=explicit symbolic]
            table[meta=label] {
            x y label
            -1.0610263636876702 -0.15389716656222058 a
            -0.782386615404361 0.6984745328493864 a
            0.14882090345415183 -0.5000264374222416 b
            -0.29882446997827505 0.360139369235527 b
            0.5449460113308662 0.04248077090154111 b
            -0.9217804637041578 0.5005564758144078 a
            0.12027829017667663 -0.5471348906278971 b
            -0.3758617102279138 -0.28765189507238775 b
            0.611549877831023 0.010841832929831166 b
            -0.8313628603905475 -0.47014310329065173 a
            0.3263934382759053 -1.0615109308715749 a
            0.24752210106481154 -0.399673330192749 b
            -0.46442060911006094 0.19493436254064422 b
            0.11866961725729615 0.5210316974354672 b
            0.024815743406605556 0.4117637875682101 b
            -0.30745057965601913 0.49814279810924494 b
            0.5722369257424879 0.844126010975173 a
            -0.7671847298508515 -0.5698114118131141 a
            0.40372601803062863 0.354287743693164 b
            -0.5906447712407059 0.8138404681611342 a
            -0.26776268616709464 -0.33260311031886824 b
            0.822997123736415 -0.6995297876436898 a
            0.1400497834450394 0.4199751044799721 b
            -0.963366698152191 -0.03948621165992599 a
            -0.204716315782816 -0.5225152508187013 b
            0.15613978129513945 -0.9696950481751906 a
            -0.2886959380114567 0.9353806632868858 a
            0.43401973115152725 -0.18871910423939253 b
            -0.7700045002676029 0.6395323128422681 a
            0.3095629137618663 0.41848177742396175 b
            -0.29893576852565845 -0.8944083328658519 a
            -0.4260166113458236 -0.9303317535102946 a
            0.895319529614891 -0.18306256365611512 a
            -0.22979036264392125 0.4169276150097926 b
            0.7812159524443781 0.4814553521237861 a
            0.37371525864741983 0.8860231045001117 a
            0.948566783056229 0.17821913305939677 a
            0.3738675295368097 0.31537831425564217 b
            -0.8410938411144048 0.3055018715247564 a
            -0.4574550965330939 0.16765028381286462 b
            -0.5838042088929276 0.8852438644760242 a
            -0.29142018055539914 0.4750223348655942 b
            0.024500977896663164 -0.5413084958864242 b
            0.24449875497823154 -0.35387356685185367 b
            -0.4090360265483381 0.09840583852758958 b
            -0.8088896635402897 -0.6776960420042523 a
            0.31114217355311513 0.9461702758603544 a
            0.2762915786250305 0.4968918235313839 b
            -0.6361832609141203 -0.7754447275488625 a
            0.6618218274178876 0.7275219893418913 a
            -0.41435441020215397 0.9125259769192932 a
            0.950524326653362 -0.10020695528606892 a
            0.07142188773241663 0.35894586443582754 b
            -0.4004716167303563 0.3542735258356427 b
            0.4824441548173157 0.14755057269842758 b
            0.45887366176630656 -0.08440423416467453 b
            0.7770216941864583 -0.5499420082330226 a
            0.4015209530871875 -0.8469575619447526 a
            0.5166877566195711 0.1722242570687679 b
            -0.09445680194448895 -0.4799201162199266 b
            -0.1077575016806574 0.9996565054027198 a
            0.9725296557558222 0.3617496469625017 a
            -0.5000889894740868 -0.1492470398397791 b
            -0.004610703713167713 0.9535537378908671 a
            -0.06428666434670804 0.5600223040904511 b
            -0.4863279318628619 -0.8405446083929345 a
            -0.4114266044428342 0.2611567246668511 b
            -0.1979250215071731 -0.9228022397927792 a
            -0.1079994793537293 -1.0275480992790167 a
            0.48955451497560104 -0.24158352430796518 b
            0.6209590722600898 0.7713573419257248 a
            -1.0219173999839042 0.1472427980584149 a
            0.4830578744909668 0.09227007210003088 b
            -0.06576079230263701 -0.42333076216420273 b
            1.070381363127673 -0.002872161269562359 a
            -0.9554476108128482 0.19671896392265767 a
            0.22449423051730816 1.0348862273224437 a
            0.05784439089294964 -0.9106072983607293 a
            -0.34636877674986416 -0.37999105408622136 b
            0.3620638792261427 -0.20874906369164314 b
            -0.19851192297444253 0.5089504025606637 b
            -0.3323341085168147 -0.38675750947780013 b
            0.7138055172738463 -0.8049016405638789 a
            -0.9963608436640177 -0.384217599373186 a
            -0.43378842222949393 -0.14775604063297795 b
            0.07007163922042997 0.5317242581196093 b
            -1.055240047881167 -0.24004451461932882 a
            0.03007937212575129 -0.5305678438196395 b
            1.0681098232552533 0.2642761642445952 a
            0.4727271421099317 -0.28057701760386394 b
            -0.3669995636736226 -0.2496942616506386 b
            -0.2544373329528893 0.9113957285395874 a
            0.2821838102686943 -0.4217371629887876 b
            -0.42724643244065447 0.0452537706274396 b
            0.42095757307566517 -0.8531779214930779 a
            0.42240479465806274 -0.15541441933898634 b
            0.7928667753890993 0.5655073040728117 a
            0.9540685815361005 -0.4083971651940238 a
            -0.4545952709763834 0.03497071653294859 b
            0.9022194918214991 -0.4623603237814011 a
                };
        \end{axis}
	\end{tikzpicture}
	\caption{Spectral clustering}
	\label{fig:spectral}
\end{figure}
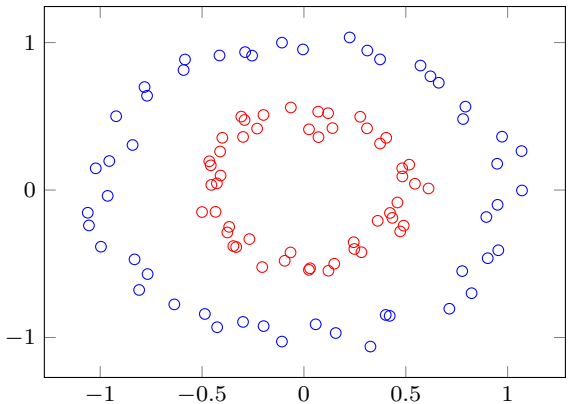

K-Means and similar methods operate based on the idea of a centroid representing an entire cluster, which is the mean of the cluster. However, there are situations in which this approach might not perform as expected, such as the synthetic set presented in Fig. \ref{fig:spectral}. In such occasions, rather than using density-based methods, an embedding named \textbf{Spectral Embedding} is used to transform the feature space based on a similarity metric to a new space which can be clustered as expected by K-Means. The similarity metric is usually obtained either using the RBF kernel or the adjacency matrix of a nearest-neighbors graph. This clustering is referred to as spectral clustering, which essentially changes the feature space of the original data. Maggioni et al. \cite{maggioni19learning} and Little et al. \cite{little2015multiscale} have proposed methods which estimate the number of clusters using spectral embedding and spectral clustering.

\subsection{Silhouette and Elbow methods}
Methods such as the elbow method and silhouette method have been employed to estimate an optimal number of clusters. These measures require sequential runs of algorithms such as K-Means and K-Means++ in order to evaluate each run and select the most appropriate one.
Silhouette is a clustering evaluation metric which can be used on any clustering output. The equation for computing the silhouette score for any given set of data $X_{n \times d}$ and their clustering labels $L_{n \times 1}$ is presented in Eq. (\ref{eq:silhouette}).
\begin{equation}
   \label{eq:silhouette}
   Silhouette(X,L) = \frac{1}{n} \sum_{i=1}^{n}{s(x_i, l_i)}, 
\end{equation}
where
\begin{equation*}
   s(x_i, l_i) =
   \begin{dcases}
    \frac{b(x_i, l_i) - a(x_i, l_i)}{max\{b(x_i,l_i),a(x_i,l_i)\}}, & \text{if } |C_{l_i}| > 1,\\
    0,              & \text{otherwise}
    \end{dcases}
\end{equation*}
\begin{equation*}
   a(x_i, l_i) = \frac{1}{|C_{l_i}| - 1} \sum\limits_{x \in C_{l_i} \setminus \{x_i\}}{\| x_i - x \|} ,
\end{equation*}
\begin{equation*}
   b(x_i, l_i) = \min_{t \in \{1,2,...,K\} \setminus \{l_i\}}{ \frac{1}{|C_{t}|} \sum\limits_{x \in C_{t}}{\| x_i - x \|}} ,
\end{equation*}
The average silhouette score is in the range $[-1,1]$, and higher silhouette scores represent better clustering. In the silhouette method, an algorithm such as K-Means or K-Means++ is run, usually starting at $K=2$ and it continues up to a specific number, which in practical cases may go up to the order of $n$. Afterwards, the optimum clustering and therefore the number of clusters is the one which produced results with the highest silhouette score.

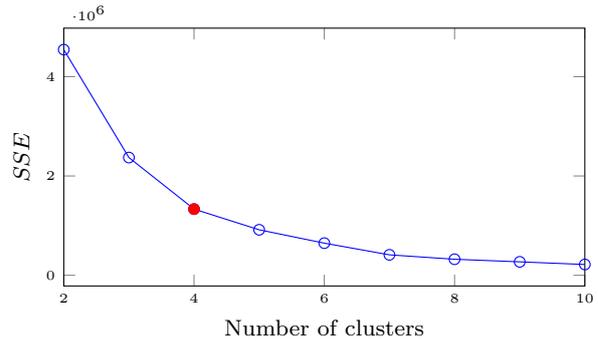
\begin{figure}
    \centering
    \begin{tikzpicture}
		\begin{axis}[
    		width=0.485\textwidth,
    		height=5cm,
    		ylabel near ticks,
    		xlabel={Number of clusters},
    		ylabel={$SSE$},
    		xmin=2, xmax=10,
    		tick label style={font=\tiny},
		]
		
		\addplot[
		color=blue,
		mark=o,
		]
		coordinates {
		(2,4543749.614531862)(3,2370689.686782968)(4,1333139.2086165315)(5,916379.187153917)(6,647326.0020260847)(7,412137.50910045847)(8,323223.2470542668)(9,270954.92924153747)(10,217968.7157736366)
		};
		
		\addplot[
		color=red,
		mark=*,
		]
		coordinates {
			(4,1333139.2086165315)
		};
		
		\end{axis}
	\end{tikzpicture}
    \caption{Wine dataset SSE values over different number of clusters, and the \textbf{elbow} point}
    \label{fig:iriselbow}
\end{figure}

In the elbow method, different $K$ values are tested and each time the sum of squared errors (presented in Eq. (\ref{eq:sse})) is logged.
\begin{equation}
   \label{eq:sse}
   SSE(X,L) = \sum\limits_{i=1}^{K}{ \sum\limits_{x \in C_i}{  \|x - z_i\|_2  } } .
\end{equation}
By definition, more clusters results in a smaller SSE, and when the number of clusters equals $n$ (each cluster has only a single data point), it is self-evident that the SSE will be equal to $0$. A plot of the SSE values on a multivariate dataset with $3$ classes is presented in Fig. \ref{fig:iriselbow}.
The \textbf{elbow} point is the breaking point of the plot from which the SSE difference between two sequential number of clusters starts to get smaller and smaller. The elbow method selects the $K$ relative to this point as the estimated number of clusters.
It is obvious that when the maximum number of clusters is unknown, these methods can be very time-consuming and even impractical, as each run is costly. Another point of weakness in these methods, which will be further discussed in detail is that their results, which are contingent on appropriate convergence of K-Means at each run.

\subsection{X-Means}
X-Means \cite{pelleg2000x} has also been used in estimating an optimal number of clusters as it runs K-Means on each cluster’s data points separately, and stops when a stopping criterion is met. This method explores the space of cluster locations and attempts to optimize the number of clusters using the Bayesian Information Criterion (BIC) or other similar measures.

\section{Proposed Method}
\label{sec:method}
As previously stated, K-Means is a clustering method which depends on the initial set of cluster centers, as well as the number of clusters. In this paper, we propose \textbf{DISCERN} (\textbf{Di}versity-based \textbf{S}election of \textbf{C}entroids for k-\textbf{E}stimation and \textbf{R}apid \textbf{N}on-stochastic clustering) as a centroid initialization method, somewhat similar to K-Means++, which selects a number of data points from the original dataset as the initial centroids and helps increase the probability of K-Means reaching better results without relying on stochastic measures. Moreover, this approach can estimate a suitable number of clusters based on the diversity of the data points it chooses. The idea behind this approach is to be able to both select the most diverse data points possible as centroids, and estimate the number of clusters while doing so.
This section is divided into three subsections: Similarity pre-computation, Diversity-Based selection and Estimation of the number of clusters.

\subsection{Similarity pre-computation}
DISCERN operates based on point-by-point similarity and starts by pre-computing a similarity matrix, $S=[s_{ij}] \in \realR^{n \times n}$:
\begin{equation}
\label{eq:similaritymatrix}
s_{ij} = cos(\theta(x_i,x_j)), \quad i,j \in \{ 1,2,...,n \}.
\end{equation}
This matrix is computed using cosine similarity with a complexity of $\bigO(dn^2)$, and requires storing the values, which yields a space complexity of $\bigO(n^2)$.

\subsection{Diversity-Based selection}
DISCERN selects the initial centroids by firstly choosing the two most diverse data points possible. This is equivalent to finding the minimum similarity in the matrix $S$ and selecting the corresponding data points, which will be the first two centroids. Therefore:
\begin{equation}
    \label{eq:discern1}
    (r_1,r_2) = \operatorname*{argmin}_{(i,j) \in \{1,2,...,n\} \times \{1,2,...,n\}}{(s_{ij})}.
\end{equation}
As a result, $r_1$ and $r_2$ are the indices of the data points which will be used as the first two centroids. Now DISCERN moves on to an iterative approach. This approach selects one data point per iteration which is expected to be the most diverse data point from the ones which are already selected. In other words, $r_3$ is the index of a data point, excluding $r_1$ and $r_2$, and this data point is different from both. In order to achieve this, a sub-matrix of $S$ named $S_3 \in \realR^{2 \times n}$ is created:
\begin{equation*}
    (S_3)_{ij} = s_{ij}, \quad i \in \{r_1,r_2\}, j \in \{1,2,...,n\} \setminus \{r_1,r_2\},
\end{equation*}
\begin{equation*}
    (S_3)_{ij} = \delta_{ij} + 1, \quad i,j \in \{r_1,r_2\}.
\end{equation*}
The matrix $S_3$ holds the similarity between the selected data points $x_{r_1}$ and $x_{r_2}$, compared to the rest. An objective is introduced which will turn this sub-matrix into a vector, and will express overall similarity. This objective is adjusted specifically (using trial and error) for expressing diversity in a way so that both the centroids and the number of clusters are selected in the best possible way.
This vector at iteration $l$, which is denoted as $p_l$, where $l \geq 3$, is generated using the sub-matrix $S_l$. The vector $p_3$ is defined as:
\begin{equation}
    \label{eq:p3computation}
    p_3=(M_3) \circ(M_3)\circ m_3 \circ (M_3 - m_3),
\end{equation}
where
\begin{equation*}
    M_3=[\max((S_3)_1),\max((S_3)_2),...,\max((S_3)_n)],
\end{equation*}
\begin{equation*}
    m_3=[\min((S_3)_1),\min((S_3)_2),...,\min((S_3)_n)],
\end{equation*}
and $(S_3)_j$ is the $j$ -th column of $S_3$.

After $p_3$ is constructed, $r_3$ is simply selected by finding the data point with the least value:
\begin{equation}
r_3 = \operatorname*{argmin}(p_3) .
\end{equation}

Assuming that the method has proceeded up to step $\ell-1$, step $\ell$ requires $S_\ell$:
\begin{equation}
\label{eq:slcomputation}
    (S_\ell)_{ij}= S_{ij},
\end{equation}
\begin{equation*}
    i \in \{r_1,r_2,...,r_{\ell-1}\}, \quad j \in \{1,2,...,n\}\setminus\{r_1,r_2,...,r_{\ell-1}\},
\end{equation*}
and
\begin{equation*}
    (S_\ell)_{ij}= \delta_{ij} + 1 \quad
i,j \in \{r_1,r_2,...,r_{\ell-1}\}.
\end{equation*}

Following that, $p_\ell$ is computed:
\begin{equation}
    \label{eq:plcomputation}
    p_\ell=(M_\ell) \circ(M_\ell)\circ m_\ell \circ (M_\ell - m_\ell),
\end{equation}
where
\begin{equation*}
    M_\ell=[\max((S_\ell)_1),\max((S_\ell)_2),...,\max((S_\ell)_n)],
\end{equation*}
\begin{equation*}
    m_\ell=[\min((S_\ell)_1),\min((S_\ell)_2),...,\min((S_\ell)_n)],
\end{equation*}
and $(S_\ell)_j$ is the $j$ -th column of $S_\ell$.

Afterwards, $r_\ell$, or the $\ell$-th centroid is selected:
\begin{equation}
r_\ell =\operatorname*{argmin}(p_\ell).
\end{equation}

By minimizing the vector $p_\ell$, the algorithm is set to minimize the maximum and minimum similarities of any given data point with the selected cluster centers, as well as the range of similarities.

\subsection{Estimation of the number of clusters}

\begin{figure*}
	\centering
	\begin{tabular}{@{}ccc@{}}
		\subfloat[Iris]{
			
			\begin{tikzpicture}
			\begin{axis}[
			width=0.3\textwidth,
			ylabel near ticks,
			xlabel={Number of centroids},
			ylabel={$R(x)$},
			xmin=0, xmax=24,
			tick label style={font=\tiny},
			]
			
			\addplot[
			color=blue,
			]
			coordinates {
		(0,0.0)(1,0.0)(2,0.02215682037950287)(3,0.03994210543564494)(4,0.040490962999090115)(5,0.04435079539276918)(6,0.0445684560555513)(7,0.04571661224777355)(8,0.04607245730660508)(9,0.04625433072403102)(10,0.04702882456113127)(11,0.047210913733982035)(12,0.04751706990766387)(13,0.04774688760724033)(14,0.047794066525014584)(15,0.04809336483831379)(16,0.048135741384190335)(17,0.04824032696293407)(18,0.04889114042857632)(19,0.04904388955822791)(20,0.049086971637883905)(21,0.049273176392985805)(22,0.05013322005011518)(23,0.050337218410325325)(24,0.050463752886221985)

			};
			
			\addplot[
			color=red,
			mark=*,
			]
			coordinates {
				(3,0.03994210543564494)
			};
			
			\end{axis}
			\end{tikzpicture}
			
		}&
		\subfloat[Wine]{
			
			\begin{tikzpicture}
			\begin{axis}[
			width=0.3\textwidth,
			ylabel near ticks,
			xlabel={Number of centroids},
			ylabel={$R(x)$},
			xmin=0, xmax=52,
			tick label style={font=\tiny},
			]
			
			\addplot[
			color=blue,
			]
			coordinates {
				(0,0.0)(1,0.0)(2,7.335095641630045e-05)(3,0.0038552407554799367)(4,0.0038711785286078184)(5,0.003904997225690766)(6,0.003938789029473983)(7,0.003957155310409406)(8,0.00396045319319544)(9,0.003992594310423818)(10,0.00402085297976933)(11,0.004056103469037317)(12,0.004101852337565521)(13,0.004211075150896148)(14,0.004296370089066186)(15,0.004303907415657373)(16,0.004325444809906198)(17,0.00433978089434897)(18,0.004397340114760783)(19,0.004408699390816536)(20,0.004459216780128128)(21,0.004463993469450517)(22,0.004556502947760087)(23,0.004587857389086826)(24,0.004605741139228379)(25,0.004697019237358564)(26,0.004742987218848876)(27,0.004755217143986037)(28,0.0048006294348140926)(29,0.004881650251070045)(30,0.004891934541140437)(31,0.004894382517367673)(32,0.00495233800590813)(33,0.0049818212016465415)(34,0.0050180946752593)(35,0.005021674910643052)(36,0.005066015142387981)(37,0.005171663473454696)(38,0.005195262043130372)(39,0.005279305707817348)(40,0.005292422671042886)(41,0.005410159373947114)(42,0.005434325294373017)(43,0.005491919928633943)(44,0.0054941005087496865)(45,0.005502779272231725)(46,0.005508458818623053)(47,0.005522362726266571)(48,0.005572971854659427)(49,0.005581783246624328)(50,0.00567940158230461)(51,0.00570341159588797)(52,0.0057735085310322435)
			};
			
			\addplot[
			color=red,
			mark=*,
			]
			coordinates {
				(3,0.0038552407554799367)
			};
			
			\end{axis}
			\end{tikzpicture}
			
		}&
		\subfloat[FEI Face Dataset]{
			
			\begin{tikzpicture}
			\begin{axis}[
			width=0.3\textwidth,
			ylabel near ticks,
			xlabel={Number of centroids},
			ylabel={$R(x)$},
			xmin=0, xmax=218,
			tick label style={font=\tiny},
			]
			
			\addplot[
			color=blue,
			]
			coordinates {
				(0,0)(1,0)(2,1.27E-06)(3,0.001783178)(4,0.003939411)(5,0.007190639)(6,0.010646449)(7,0.011252816)(8,0.012693156)(9,0.013884351)(10,0.01728448)(11,0.017846191)(12,0.020893329)(13,0.021110227)(14,0.024473475)(15,0.024783887)(16,0.029124567)(17,0.031154226)(18,0.0326839)(19,0.03297305)(20,0.03314286)(21,0.035429924)(22,0.035945162)(23,0.03686513)(24,0.03749668)(25,0.03760834)(26,0.038619946)(27,0.040295564)(28,0.041816749)(29,0.042206276)(30,0.043039977)(31,0.043108639)(32,0.043392814)(33,0.043820221)(34,0.044223371)(35,0.044648255)(36,0.04507389)(37,0.045076652)(38,0.045541901)(39,0.046369377)(40,0.046380156)(41,0.046509577)(42,0.046628645)(43,0.047076921)(44,0.047814636)(45,0.048062565)(46,0.049554458)(47,0.049894345)(48,0.050598206)(49,0.051388808)(50,0.051907388)(51,0.05193165)(52,0.052267123)(53,0.053004307)(54,0.053182623)(55,0.05322607)(56,0.053374758)(57,0.053695883)(58,0.054483376)(59,0.05448572)(60,0.054884794)(61,0.055612876)(62,0.056094656)(63,0.056384963)(64,0.056837762)(65,0.056894655)(66,0.057105738)(67,0.057339534)(68,0.05783115)(69,0.058181871)(70,0.058585553)(71,0.058882734)(72,0.059351055)(73,0.05993731)(74,0.060409285)(75,0.060503317)(76,0.061195339)(77,0.061438692)(78,0.06144157)(79,0.062821459)(80,0.063047109)(81,0.063057108)(82,0.063227977)(83,0.063264812)(84,0.06383008)(85,0.064247959)(86,0.064345362)(87,0.065103427)(88,0.065109768)(89,0.065591768)(90,0.065677743)(91,0.065769589)(92,0.066382277)(93,0.066655204)(94,0.066860083)(95,0.067695815)(96,0.067720284)(97,0.067960243)(98,0.068548384)(99,0.068680027)(100,0.068822373)(101,0.068937923)(102,0.069380982)(103,0.069709054)(104,0.070183061)(105,0.071024529)(106,0.071250101)(107,0.071304456)(108,0.071320429)(109,0.072185786)(110,0.072601606)(111,0.072745901)(112,0.073447832)(113,0.074158467)(114,0.074275552)(115,0.074316484)(116,0.075373367)(117,0.075688487)(118,0.075802071)(119,0.077000889)(120,0.077131895)(121,0.077625275)(122,0.077766987)(123,0.07784692)(124,0.078219485)(125,0.078304828)(126,0.079483917)(127,0.079671577)(128,0.079952535)(129,0.080033826)(130,0.080524243)(131,0.080666158)(132,0.080929066)(133,0.081245311)(134,0.08150062)(135,0.081968477)(136,0.082193038)(137,0.082648982)(138,0.082953891)(139,0.083407735)(140,0.084334109)(141,0.08487278)(142,0.084981073)(143,0.085840235)(144,0.08646293)(145,0.086709576)(146,0.08727776)(147,0.087728888)(148,0.087734756)(149,0.088704191)(150,0.088832677)(151,0.08910105)(152,0.090198668)(153,0.090289492)(154,0.09103121)(155,0.09156522)(156,0.092062073)(157,0.092429086)(158,0.092640314)(159,0.093568996)(160,0.094114189)(161,0.09450056)(162,0.09513047)(163,0.095475819)(164,0.095549578)(165,0.095712742)(166,0.095937912)(167,0.097257461)(168,0.097858431)(169,0.098178535)(170,0.098400222)(171,0.099028548)(172,0.099721682)(173,0.09991583)(174,0.100684251)(175,0.100904964)(176,0.101188874)(177,0.101207374)(178,0.101276244)(179,0.102915767)(180,0.103119145)(181,0.103427218)(182,0.105550444)(183,0.10670085)(184,0.10708677)(185,0.107937744)(186,0.108142688)(187,0.108193681)(188,0.109312904)(189,0.110296534)(190,0.112007968)(191,0.11268857)(192,0.114770251)(193,0.115233674)(194,0.116864645)(195,0.117653558)(196,0.121324071)(197,0.123229318)(198,0.126113848)(199,0.135020577)(200,0.137976903)(201,0.147215014)(202,0.150316915)(203,0.150895735)(204,0.15691528)(205,0.15829441)(206,0.163520535)(207,0.164898703)(208,0.165929522)(209,0.166032323)(210,0.167990949)(211,0.170501392)(212,0.171386465)(213,0.172464095)(214,0.173173113)(215,0.173572412)(216,0.174989757)(217,0.177243939)(218,0.177482242)
			};
			
			\addplot[
			color=red,
			mark=*,
			]
			coordinates {
				(199,0.135020577)
			};
			
			\end{axis}
			\end{tikzpicture}
			
		}\\
	\end{tabular}
	
	\caption{The function $R$ and the target number of clusters (red point) of the dataset}
	\label{fig2:rfunction}
\end{figure*}
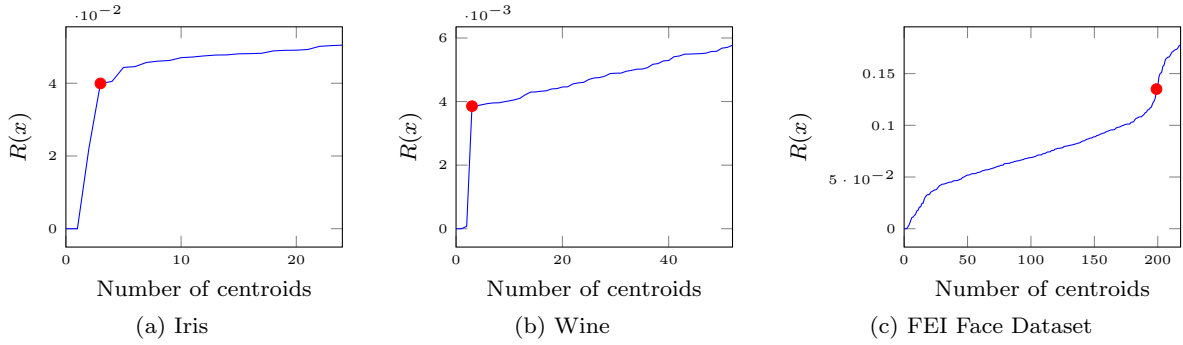
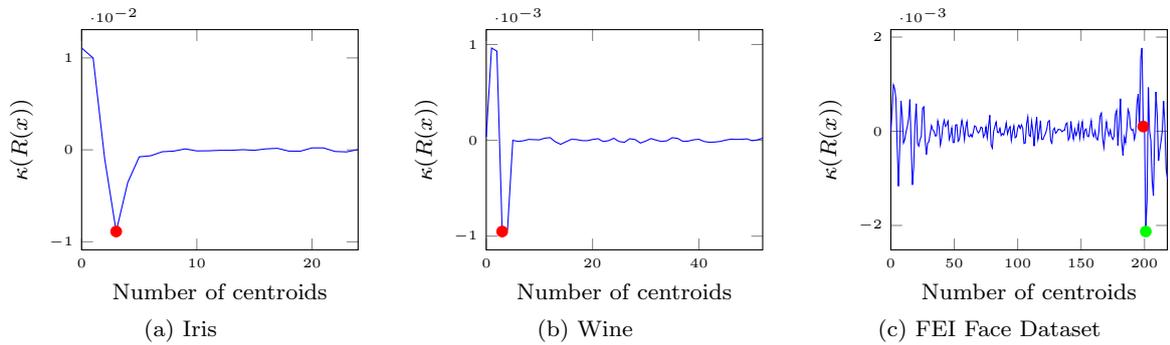
\begin{figure*}
	\centering
	\begin{tabular}{@{}ccc@{}}
		\subfloat[Iris]{
			
			\begin{tikzpicture}
			\begin{axis}[
			width=0.3\textwidth,
			ylabel near ticks,
			xlabel={Number of centroids},
			ylabel={$\kappa(R(x))$},
			xmin=0, xmax=24,
			tick label style={font=\tiny},
			]
			
			\addplot[
			color=blue,
			]
			coordinates {
			(0,0.011078410189751435)(1,0.00998368833785764)(2,-0.0009550979818677118)(3,-0.008882234215663254)(4,-0.0035641364128104346)(5,-0.0007607135326807094)(6,-0.0006433725012832158)(7,-0.00020702441907670266)(8,-0.0001369084842871926)(9,0.00010471609750687582)(10,-0.00011703043684008644)(11,-0.00010515227477321691)(12,-5.281217660625334e-05)(13,-4.737415918312644e-05)(14,1.6169559728347456e-05)(15,-4.9878774429695935e-05)(16,0.00010343104546484962)(17,0.00016415008254266767)(18,-0.00013989192489584362)(19,-0.0001435689380692913)(20,0.00021260429653949925)(21,0.0002086887099811047)(22,-0.00017892881806338864)(23,-0.00022470061206152317)(24,2.7941119906847068e-05)
			};
			
			\addplot[
			color=red,
			mark=*,
			]
			coordinates {
(3,-0.008882234215663254)			};
			
			\end{axis}
			\end{tikzpicture}
			
		}&
		\subfloat[Wine]{
			
			\begin{tikzpicture}
			\begin{axis}[
			width=0.3\textwidth,
			ylabel near ticks,
			xlabel={Number of centroids},
			ylabel={$\kappa(R(x))$},
			xmin=0, xmax=52,
			tick label style={font=\tiny},
			]
			
			\addplot[
			color=blue,
			]
			coordinates {
				(0,3.667547820815023e-05)(1,0.000963810186925366)(2,0.0009311139643003665)(3,-0.0009513659255547728)(4,-0.0009325542669655643)(5,6.00403625923523e-07)(6,-1.1486584274458633e-05)(7,-4.179771175321385e-06)(8,9.683905708547537e-06)(9,7.017539640171357e-06)(10,5.149892797785845e-06)(11,2.286563075507589e-05)(12,2.8379598170529348e-05)(13,-1.5534854053978418e-05)(14,-4.13607575314981e-05)(15,-1.4239696512892975e-05)(16,1.0705145998476966e-05)(17,8.261254427978786e-06)(18,-2.5046598673487333e-06)(19,-3.4061044535056103e-06)(20,8.852375556003841e-06)(21,1.7142460189739508e-05)(22,-1.2011993971807552e-05)(23,-3.6755178378010586e-06)(24,2.200197193973304e-05)(25,-1.2740985321068292e-05)(26,-1.9900965888543588e-05)(27,1.705880009287915e-05)(28,8.415722539833957e-06)(29,-2.8425210107731424e-05)(30,-7.725410389193245e-06)(31,1.867660446975652e-05)(32,1.3383011420323379e-06)(33,-1.1896243801300052e-05)(34,-4.459050552966352e-06)(35,2.7533713430073263e-05)(36,2.033160823190563e-05)(37,-1.0586582045930935e-05)(38,-8.021568172614856e-06)(39,5.802857921236141e-06)(40,1.1185498782581959e-05)(41,-1.2273277768057196e-05)(42,-2.053185218689607e-05)(43,-1.7725302748511398e-05)(44,-1.1354226125323735e-05)(45,2.1810276090971395e-06)(46,1.25386815389486e-05)(47,9.959266565184228e-06)(48,1.047917288832725e-05)(49,1.5551957160410858e-05)(50,-3.0806947122969277e-06)(51,4.641380240811691e-06)(52,2.3043460610617423e-05)
			};
			
			\addplot[
			color=red,
			mark=*,
			]
			coordinates {
				(3,-0.0009513659255547728)
			};
			
			\end{axis}
			\end{tikzpicture}
			
		}&
		\subfloat[FEI Face Dataset]{
			
			\begin{tikzpicture}
			\begin{axis}[
			width=0.3\textwidth,
			ylabel near ticks,
			xlabel={Number of centroids},
			ylabel={$\kappa(R(x))$},
			xmin=0, xmax=218,
			tick label style={font=\tiny},
			]
			
			\addplot[
			color=blue,
			]
			coordinates {
				(0,6.33E-07)(1,0.000445794)(2,0.000984219)(3,0.000906066)(4,0.000692216)(5,-0.000336315)(6,-0.001165076)(7,-0.00035766)(8,0.000636152)(9,0.000332574)(10,-0.000245617)(11,-0.00017445)(12,-7.18E-06)(13,0.000102405)(14,0.000267735)(15,0.000674164)(16,-0.000272935)(17,-0.001137873)(18,-0.000775092)(19,0.000159513)(20,0.000585834)(21,-0.000255417)(22,-0.000312696)(23,-0.000172999)(24,-0.000107063)(25,0.000486004)(26,0.000518383)(27,-0.000194127)(28,-0.000493393)(29,-0.000252087)(30,-0.000217598)(31,-4.77E-05)(32,0.00011943)(33,2.91E-05)(34,4.99E-06)(35,-9.99E-05)(36,-9.56E-05)(37,0.000216082)(38,9.26E-05)(39,-0.000288131)(40,-0.000147442)(41,0.000106786)(42,0.000234376)(43,0.000104575)(44,0.000138457)(45,0.000211534)(46,-0.000174018)(47,-8.43E-05)(48,6.64E-05)(49,-0.000237905)(50,-0.000237362)(51,0.000132454)(52,0.000138941)(53,-0.000212723)(54,-0.000180841)(55,6.20E-05)(56,0.000229121)(57,8.00E-05)(58,-0.0001768)(59,8.43E-05)(60,0.000202111)(61,-8.88E-05)(62,-0.000116689)(63,-6.56E-05)(64,-0.000118782)(65,-1.62E-05)(66,0.000114359)(67,9.94E-05)(68,7.25E-06)(69,-3.54E-05)(70,2.77E-06)(71,8.84E-05)(72,7.32E-05)(73,-0.000122142)(74,-6.80E-05)(75,9.23E-05)(76,-0.000134956)(77,0.000111848)(78,0.000339827)(79,-0.000286779)(80,-0.000356168)(81,-6.99E-06)(82,0.000105309)(83,0.000193861)(84,-2.17E-05)(85,-3.19E-05)(86,6.23E-05)(87,-9.18E-05)(88,-4.91E-05)(89,-7.76E-05)(90,3.41E-05)(91,0.000176948)(92,-5.67E-05)(93,3.87E-05)(94,9.56E-05)(95,-0.000194045)(96,-8.03E-06)(97,0.000113839)(98,-0.000138528)(99,-0.000115472)(100,7.12E-05)(101,0.000128309)(102,6.09E-05)(103,0.000136086)(104,6.62E-05)(105,-0.000258887)(106,-0.000249178)(107,0.000150351)(108,0.000302712)(109,-8.03E-05)(110,-0.000108737)(111,0.000213113)(112,-4.63E-06)(113,-0.000313637)(114,6.75E-05)(115,0.000303496)(116,-0.000167277)(117,-1.49E-05)(118,0.00022528)(119,-0.000172004)(120,-0.000173683)(121,-0.000100685)(122,-4.56E-05)(123,5.91E-05)(124,0.000202983)(125,0.00022721)(126,-0.000198953)(127,-0.000251125)(128,2.58E-05)(129,6.75E-05)(130,-4.17E-05)(131,-1.33E-05)(132,4.17E-05)(133,3.60E-05)(134,3.02E-05)(135,-1.07E-05)(136,1.71E-05)(137,1.96E-05)(138,0.000154841)(139,0.000176573)(140,-0.000183313)(141,-0.000124398)(142,0.000208723)(143,-2.45E-05)(144,-0.000166757)(145,3.75E-05)(146,-8.95E-05)(147,-1.10E-05)(148,0.000160231)(149,-0.000144611)(150,6.70E-05)(151,0.000197896)(152,-0.000133362)(153,2.18E-05)(154,4.96E-05)(155,-0.000102966)(156,-0.000113156)(157,6.90E-05)(158,0.000223908)(159,-5.21E-05)(160,-0.000114398)(161,1.09E-05)(162,-0.000149293)(163,-0.000184584)(164,-7.69E-06)(165,0.000326949)(166,0.000383046)(167,-0.000155911)(168,-0.000344682)(169,-1.78E-05)(170,0.000194917)(171,9.32E-06)(172,-8.97E-05)(173,2.55E-05)(174,-0.000114486)(175,-0.000171681)(176,-0.000104313)(177,0.000351496)(178,0.000438882)(179,-0.000299235)(180,0.0001471)(181,0.000690544)(182,-0.000223742)(183,-0.000509184)(184,-0.000120102)(185,-0.000245239)(186,2.86E-05)(187,0.000461729)(188,0.000381211)(189,7.23E-05)(190,1.68E-05)(191,3.83E-05)(192,-0.000166972)(193,-3.13E-05)(194,0.000591257)(195,0.000788963)(196,8.26E-05)(197,0.001553861)(198,0.001768227)(199,0.000100789)(200,0.000119232)(201,-0.002128308)(202,-0.001435404)(203,0.000929473)(204,1.72E-06)(205,-0.000198592)(206,-0.00104905)(207,-0.001367665)(208,-8.69E-05)(209,0.000833861)(210,0.00033352)(211,-0.000626589)(212,-0.000402217)(213,-0.000213596)(214,7.50E-06)(215,0.000640802)(216,0.00016896)(217,-0.000798728)(218,-0.001007939)
			};
		
			\addplot[
			color=red,
			mark=*,
			]
			coordinates {
				(199,0.000100789)
			};
			
			\addplot[
			color=green,
			mark=*,
			]
			coordinates {
				(201,-0.002128308)
			};
			
			\end{axis}
			\end{tikzpicture}
			
		}\\
	\end{tabular}
	\caption{The curvature of $R$, the target number of clusters (red points) and the estimated number of clusters (if different from red, green point).}
	\label{fig3:kappa}
\end{figure*}

The vector $p_\ell$ (Eq. (\ref{eq:plcomputation})) aims to minimize similarity between centroids and was carefully adjusted so that its values could help estimate the number of clusters. To that end, the function $R$ is defined as follows:
\begin{equation}
    \label{eq:membershipfunction}
    R:\{1,2,...,n\} \rightarrow [0,1],
\end{equation}
where
\begin{equation*}
    R(\ell) = 
   \begin{dcases}
    0,              & \text{if } \ell=1,2,\\
    \min(p_\ell), & \text{if } \ell=3,4,...,n.
    \end{dcases}
\end{equation*}

Many different variations of the objective function which generates the $p_\ell$ vectors have been previously explored, and the one presented in Eq. (\ref{eq:plcomputation}) performed best.
Instances of the function $R$ on three datasets are presented in Figure \ref{fig2:rfunction}. It is self-evident that the target $K$ is at a breaking point. We found that calculating the curvature of this function using finite differences can help detect this point which can serve as a good estimate for the number of clusters. The signed curvature of $R$ is defined as:
\begin{equation}
\label{eq:kappaR}
\kappa(R) = \frac{R^{\prime\prime}}{(1+R^{\prime 2})^\frac{3}{2}} 
\end{equation}
The results showed that an optimum $K$ is usually close to the minimum value of curvature ($\kappa(R)$). The graphs of $\kappa(R)$ on the three instances in Figure \ref{fig2:rfunction} are expressed in Figure \ref{fig3:kappa}.
The red points represent the target number of clusters, which is the same as the minimum value of the signed curvature of $R$ in Figure \ref{fig3:kappa}.a and \ref{fig3:kappa}.b. In Figure \ref{fig3:kappa}.c however, the minimum of curvature of $R$ is the green point which is different from the exact target (the target being $199$ clusters and the minimum of $\kappa(R)$ being at $201$).
The optimal number of clusters is estimated as:
\begin{equation}
\label{eq:discernkestimation}
K=\operatorname*{argmin}(\kappa(R)).
\end{equation}
The resulting centroids are:
\begin{equation}
Z = 
	\begin{bmatrix}
    x_{r_1} \\ x_{r_2} \\ ... \\ x_{r_K}
    \end{bmatrix}, \quad K\geq 2
\end{equation}
which are the initial centroids that can guide K-Means to converge to more appropriate results. We also provide the pseudo-code of DISCERN in Algorithm \ref{alg:discern}.
\begin{algorithm}
	\SetKwInOut{Input}{Input}
	\SetKwInOut{Output}{Output}
	\Input{Dataset $X \in \realR^{n \times d}$}
	\Output{Cluster assignments $L$, $K$}
	Compute similarity matrix $S_{n \times n}$\;
	$(r_1,r_2) = \operatorname{argmin}(S)$\;
	$R(1) = R(2) = 0$\;
	\For{$\ell = 3$ \KwTo $n$}
	{
		Create $S_{\ell} \in \realR^{(\ell - 1) \times n}$ according to Eq. (\ref{eq:slcomputation})\;
		$(M_l,m_l) = \max(S_{\ell}),\min(S_{\ell})$, using Eq. (\ref{eq:plcomputation})\;
		Compute $p_{\ell}$ according to according to Eq. (\ref{eq:plcomputation})\;
		$(R(\ell),r_{\ell}) = (\min(p_{\ell}),\operatorname{argmin}{(p_{\ell})})$\;
	}
	Compute $\kappa(R)$ according to Eq. (\ref{eq:kappaR})\;
	$K = \operatorname{argmin}{(\kappa(R))}$\;
	$Z = 
	\begin{bmatrix}
    x_{r_1} \\ x_{r_2} \\ ... \\ x_{r_K}
    \end{bmatrix}$\;
	$L = \text{K-Means}(X,K,Z,t)$\;
	\caption{DISCERN}
	\label{alg:discern}
\end{algorithm}

\begin{figure}
	\centering
	\includegraphics[width=0.35\textwidth, height=4.5cm]{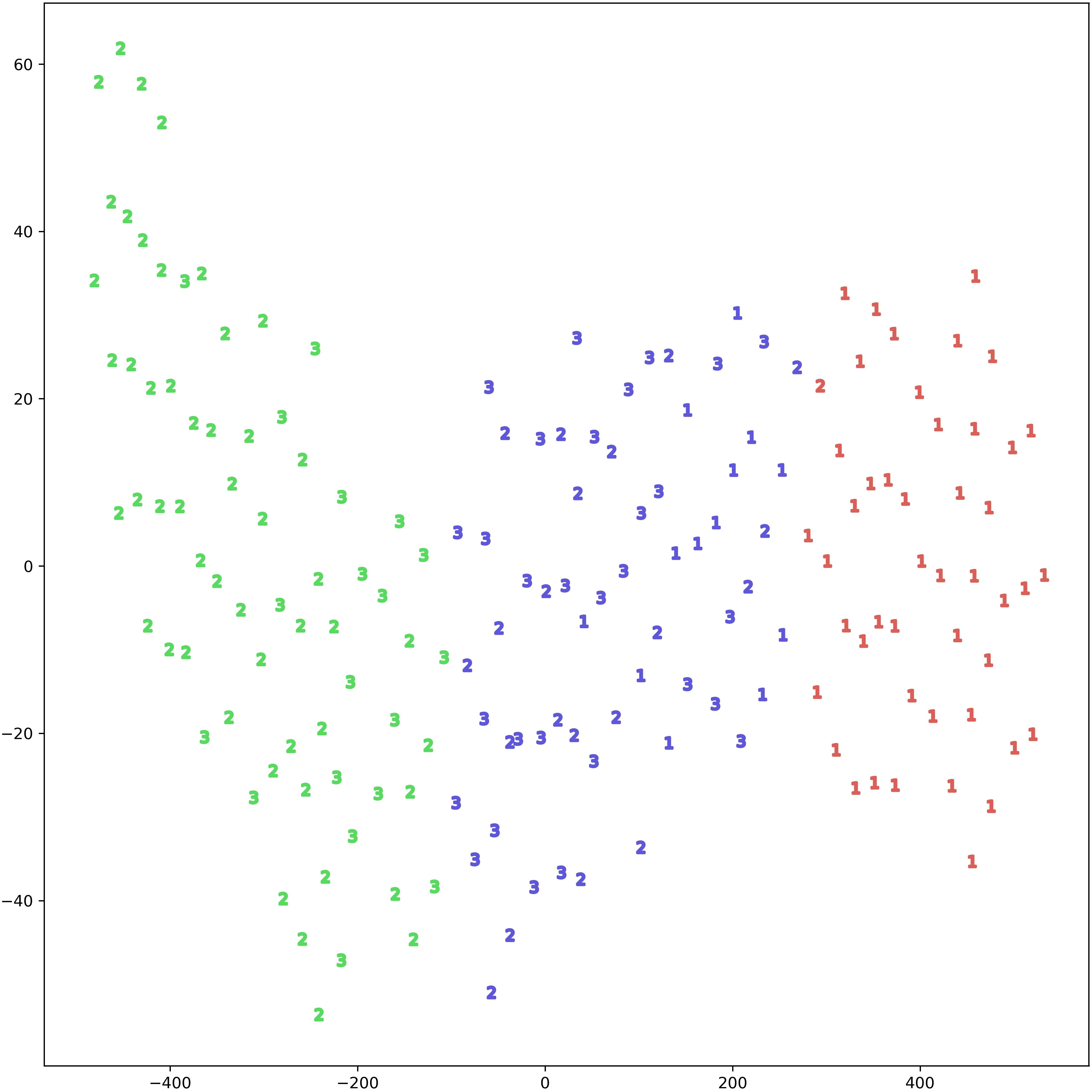}
	\caption{Wine dataset(t-SNE Visualization)}
	\label{fig:winetsne}
\end{figure}

Therefore DISCERN can estimate the number of clusters without having to run K-Means with different $K$ values and provides suitable initial centroids.
We do have to note that in order for clustering algorithms to serve as practical classifiers, the feature space is required to be in relation with what is expected. An example of such a dataset can be Wine (t-SNE visualization provided in Fig. \ref{fig:winetsne}), in which the $13$-dimensional feature space somewhat correlates with the three classes.

\section{Experiments and analysis}
\label{sec:experiments}
\begin{table*}
	\centering
	\caption{Datasets and Their Characteristics}
	\label{tab:datasets}
	\begin{tabular}{p{0.175\textwidth}p{0.1\textwidth}p{0.075\textwidth}p{0.1\textwidth}p{0.075\textwidth}p{0.1\textwidth}p{0.15\textwidth}}
		\hline\noalign{\smallskip}
		Dataset&Abbreviation&Features&Samples&Classes&Type&Embedding\\
		\noalign{\smallskip}\hline\noalign{\smallskip}
		Yale Faces\cite{belhumeur1997eigenfaces}&YALE&128&165&15&Image&FaceNet\\
		MIT-CBCL Faces\cite{weyrauch2004component}&MIT&128&59&10&Image&FaceNet\\
		GeorgiaTech Faces\cite{gatechfaces}&GA&128&750&50&Image&FaceNet\\
		AT\&T Faces\cite{samaria1994parameterisation}&ATT&128&400&40&Image&FaceNet\\
		Caltech Faces\cite{caltechfaces}&CA&128&450&31&Image&FaceNet\\
		FEI Faces\cite{feidataset}&FEI&128&400&199&Image&FaceNet\\
		ImageNette\cite{imagenettedataset}&IMG&512&9469 + 3925&10&Image&ResNet101 + PCA\\
		ImageWoof\cite{imagenettedataset}&WOOF&2048&9025 + 3929&10&Image&ResNet101\\
		Wine&WINE&13&178&3&Multivariate&Original\\
		Iris&IRIS&4&150&3&Multivariate&Original\\
		Prestige\cite{prestigedataset}&PRES&5&102&4&Multivariate&Original\\
		MFeat Fourier&MFF&76&2000&10&Multivariate&Original\\
		Wap\cite{han1998webace}&WAP&30&30&20&Text&PCA\\
		BBC News\cite{greene2006practical}&BBC&5&29392&5&Text&LSA\\
		\noalign{\smallskip}\hline
	\end{tabular}
\end{table*}

\begin{table*}
	\centering
	\caption{Estimated K comparison between the proposed approach and other measures.}
	\label{tab:kestimation}
	\begin{tabular}{p{0.075\textwidth}p{0.12\textwidth}p{0.12\textwidth}p{0.12\textwidth}p{0.12\textwidth}p{0.12\textwidth}}
		\hline\noalign{\smallskip}
		Dataset&True K&Silhouette&Elbow&X-Means&DISCERN\\
		\noalign{\smallskip}\hline\noalign{\smallskip}
		
		YALE&15&16&14&15&15\\
		MIT&10&13&11&2&12\\
		GA&50&56&32&52&54\\
		ATT&40&44&44&4&40\\
		CA&31&29&34&25&41\\
		FEI&199&216&69&2&201\\
		IMG&10&11&13&46&10\\
		WOOF&10&9&11&38&10\\
		WINE&3&2&4&10&3\\
		IRIS&3&2&4&6&3\\
		PRES&4&2&6&13&4\\
		MFF&10&2&8&20&5\\
		WAP&20&28&17&40&18\\
		BBC&5&5&6&15&6\\
		\noalign{\smallskip}\hline
	\end{tabular}
\end{table*}
In this section, we compare DISCERN with some of the methods mentioned in Sect. \ref{sec:relatedwork}.
We present the details of the experiments we conducted in order to compare each aspect of the proposed approach.
Firstly, comparison to methods that estimate the number of clusters is presented, which are: The silhouette and elbow methods and X-Means \cite{pelleg2000x}. The metric for comparison is the proximity of the estimated number of clusters to the number of ground truth classes.
Afterwards, we present the clustering performance of both K-Means and K-Means++ compared to DISCERN, using both the number of ground truth classes as the number of clusters, as well as the estimated number of clusters by DISCERN. The metrics used for this comparison are the Average Silhouette Coefficient (presented in Eq. (\ref{eq:silhouette})), Adjusted Rand Index (ARI) \cite{hubert1985comparing} and Purity (clustering accuracy).
Silhouette is an internal metric, which means it only relies on the dataset and the clustering labels, while ARI is an external one, requiring the original classification labels as well. Given:
\begin{flalign}
\label{eq:aripre}
\begin{aligned}
&T_j = \{ x_i | x_i \textnormal{ is in class j} \}\\
&n_{ij} = | C_i \cap T_j |\\
&a_i = \sum_{j=1}^{t}{n_{ij}}, \quad t \textnormal{ is the number of classes}\\
&b_j = \sum_{i=1}^{K}{n_{ij}}, \quad K \textnormal{ is the number of clusters}\\
\end{aligned}
\end{flalign}
ARI is then computed using:
\begin{equation}
    \label{eq:ari}
    \text{ARI} = \frac{\sum_{ij}{\binom{n_{ij}}{2}} - \frac{( \sum_i{\binom{a_i}{2}} \sum_j{ \binom{b_j}{2} } )}{\binom{n}{2}}}{ \frac{1}{2} ( \sum_i{\binom{a_i}{2}} + \sum_j{ \binom{b_j}{2} )  - \frac{( \sum_i{\binom{a_i}{2}} \sum_j{ \binom{b_j}{2} } )}{\binom{n}{2}}}}
\end{equation}
Purity is also an external evaluation metric, and is measured using Eq. (\ref{eq:purity}).
\begin{equation}
\label{eq:purity}
    \text{Purity} = \frac{1}{n} \sum\limits_{i=1}^{K}{\max\limits_{j}n_{ij}}
\end{equation}

We present a summary of datasets used in our experiments in Table \ref{tab:datasets}. A pre-trained FaceNet\cite{schroff2015facenet}, obtained from a GitHub repository \cite{facenetkeras} trained on \textbf{MS-Celeb-1M} \cite{guo2016ms} was used to embed facial image sets. For the two ImageNet subsets (ImageNette and ImageWoof), we used a pre-trained ResNet101 \cite{he2016deep} to embed the training and test sets. These sets were split into training and validation sets, the sizes of which are presented in the table. For experiments on these two sets, all clustering methods were trained on the training set and later evaluated using the validation sets.
Principal Component Analysis (PCA) was applied to Wap and it was reduced to $\realR^{30}$ prior to running the tests as it was better suited for clustering. Latent Semantic Analysis (LSA) was applied to the text dataset BBC News.
In the experiments, facial data are clustered using cosine distance, since it is more suitable for the latent space of FaceNet. Three other datasets, Iris, Wap and Prestige were also clustered using cosine distance, while the rest of the datasets were clustered using Euclidean distance.
The libraries used in the experiments are: PyTorch \cite{pytorch}, TensorFlow \cite{tensorflow2015}, Numpy \cite{van2011numpy}, Scikit-Learn \cite{pedregosa2011scikit}, PyClustering \cite{novikov2019pyclustering} and KEEL \cite{alcala2011keel}.

\subsection{K-Estimation Performance}

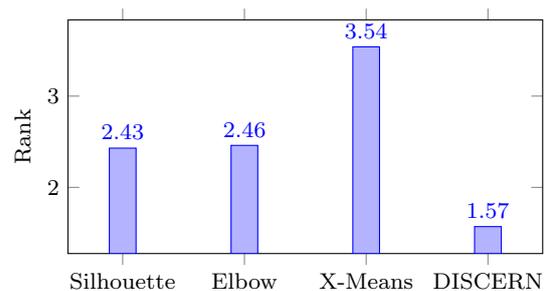
\begin{figure}[!ht]
	\centering
	\begin{tikzpicture}
	\begin{axis}[
	width=0.45\textwidth,
	height=0.20\textheight,
	ybar,
	enlargelimits=0.15,
	ylabel near ticks,
	ylabel=Rank,
	symbolic x coords={Silhouette,Elbow,X-Means,DISCERN},
	nodes near coords,
	nodes near coords align={vertical},
	]
	\addplot coordinates {(Silhouette,2.43)(Elbow,2.46)(X-Means,3.54)(DISCERN,1.57)};
	\end{axis}
	\end{tikzpicture}
	\caption{Friedman ranking test on the methods presented in Table \ref{tab:kestimation}.}
	\label{fig:kcomprank}
\end{figure}

In this subsection, we estimated an optimum $K$ for each dataset. For the other methods we used K-Means++ for estimation. The estimated $K$ and the real number of classes are presented in Table \ref{tab:kestimation}.
Each method was performed multiple times and the results were averaged, and rounded to the nearest integer.
A Friedman test was performed on these results, with the Friedman statistic being computed using chi-square with 3 degrees of freedom. The resulting rankings are presented in Figure \ref{fig:kcomprank}.
The proposed approach outranks the rest in this test and Li’s post-hoc p-value comparison of these methods to DISCERN are presented in Table \ref{tab:kestimationpvalues}. The p-value when compared against X-Means is below the $5 \%$ limit which points to significant improvement, while it is not necessarily the case for the other two.
\begin{table}[!ht]
	\centering
	\caption{K-Estim. p-value comparison to DISCERN}
	\label{tab:kestimationpvalues}
	\begin{tabular}{p{0.085\textwidth}p{0.085\textwidth}p{0.085\textwidth}}
		\hline\noalign{\smallskip}
		Silhouette&Elbow&X-Means\\
		\noalign{\smallskip}\hline\noalign{\smallskip}
		$7.9 \times 10^{-2}$&$6.8 \times 10^{-2}$&$6.2 \times 10^{-5}$\\
		\noalign{\smallskip}\hline
	\end{tabular}
\end{table}
Therefore, the significance of the improvement over the Elbow and Silhouette methods cannot be inferred with certainty. Nevertheless, DISCERN's advantage over these methods is less complexity, which will be discussed in \ref{sec:complexitycomparison}.

\subsection{Clustering Performance}
\begin{table*}
	\centering
	\caption{Comparison of clustering methods set to True K.}
	\label{tab:truekcomparison}
	\begin{tabular}{p{0.05\textwidth}|ccc|ccc|ccc}
		\hline\noalign{\smallskip}
		\multirow{2}{*}{Dataset} &
		\multicolumn{3}{c}{\textbf{ASC}} &
		\multicolumn{3}{c}{\textbf{Purity}} &
		\multicolumn{3}{c}{\textbf{ARI}} \\
		& K-Means & K-Means++ & DISCERN & K-Means & K-Means++ & DISCERN & K-Means & K-Means++ & DISCERN \\
		\noalign{\smallskip}\hline\noalign{\smallskip}
		YALE&0.544&0.581&\textbf{0.723}&0.834&0.852&\textbf{1}&0.791&0.798&\textbf{1}\\
		MIT&0.493&0.549&\textbf{0.653}&0.835&0.863&\textbf{0.983}&0.751&0.789&\textbf{0.96}\\
		GA&0.494&0.517&\textbf{0.625}&0.841&0.847&\textbf{0.96}&0.802&0.785&\textbf{0.94}\\
		ATT&0.569&0.619&\textbf{0.751}&0.851&0.883&\textbf{1}&0.807&0.846&\textbf{1}\\
		CA&0.595&0.678&\textbf{0.83}&0.898&\textbf{0.957}&0.953&0.828&0.905&\textbf{0.942}\\
		FEI&0.47&0.602&\textbf{0.79}&0.839&0.903&\textbf{0.995}&0.636&0.763&\textbf{0.985}\\
		IMG&0.126&\textbf{0.159}&\textbf{0.159}&0.728&\textbf{0.825}&\textbf{0.825}&0.614&\textbf{0.74}&\textbf{0.74}\\
		WOOF&0.163&0.169&\textbf{0.170}&0.891&\textbf{0.911}&0.892&0.836&0.862&\textbf{0.917}\\
		WINE&0.728&0.729&\textbf{0.732}&0.697&0.693&\textbf{0.702}&0.366&0.361&\textbf{0.371}\\
		IRIS&\textbf{0.752}&0.747&0.748&0.933&0.895&\textbf{0.973}&0.857&0.8&\textbf{0.922}\\
		PRES&0.146&0.143&\textbf{0.151}&0.753&0.752&\textbf{0.765}&\textbf{0.395}&0.386&0.382\\
		MFF&0.258&0.264&\textbf{0.269}&0.664&0.692&\textbf{0.731}&0.512&0.542&\textbf{0.577}\\
		WAP&0.316&0.317&\textbf{0.333}&0.635&0.614&\textbf{0.636}&0.3&0.249&\textbf{0.416}\\
		BBC&0.549&0.539&\textbf{0.557}&0.757&0.749&\textbf{0.788}&0.502&0.491&\textbf{0.55}\\
		\noalign{\smallskip}\hline
	\end{tabular}
\end{table*}
\begin{table*}
	\centering
	\caption{Comparison of clustering methods set to DISCERN K.}
	\label{tab:discernkcomparison}
	\begin{tabular}{p{0.05\textwidth}|ccc|ccc|ccc}
		\hline\noalign{\smallskip}
		\multirow{2}{*}{Dataset} &
		\multicolumn{3}{c}{\textbf{ASC}} &
		\multicolumn{3}{c}{\textbf{Purity}} &
		\multicolumn{3}{c}{\textbf{ARI}} \\
		& K-Means & K-Means++ & DISCERN & K-Means & K-Means++ & DISCERN & K-Means & K-Means++ & DISCERN \\
		\noalign{\smallskip}\hline\noalign{\smallskip}
		YALE&0.534&0.599&\textbf{0.723}&0.815&0.868&\textbf{1}&0.759&0.822&\textbf{1}\\
		MIT&0.504&0.54&\textbf{0.667}&0.878&0.917&\textbf{0.966}&0.756&0.799&\textbf{0.887}\\
		GA&0.492&0.521&\textbf{0.607}&0.863&0.882&\textbf{0.96}&0.806&0.809&\textbf{0.919}\\
		ATT&0.572&0.622&\textbf{0.751}&0.851&0.888&\textbf{1}&0.808&0.853&\textbf{1}\\
		CA&0.515&0.588&\textbf{0.775}&0.947&0.985&\textbf{1}&0.816&0.873&\textbf{0.956}\\
		FEI&0.469&0.609&\textbf{0.793}&0.843&0.909&\textbf{1}&0.634&0.769&\textbf{0.995}\\
		IMG&0.126&\textbf{0.159}&\textbf{0.159}&0.728&\textbf{0.825}&\textbf{0.825}&0.614&\textbf{0.74}&\textbf{0.74}\\
		WOOF&0.163&0.169&\textbf{0.170}&0.891&\textbf{0.911}&0.892&0.836&0.862&\textbf{0.917}\\
		WINE&0.73&0.73&\textbf{0.732}&0.697&0.695&\textbf{0.702}&0.366&0.362&\textbf{0.371}\\
		IRIS&\textbf{0.764}&0.756&0.748&0.819&0.924&\textbf{0.973}&0.681&0.847&\textbf{0.922}\\
		WAP&0.313&0.299&\textbf{0.319}&0.616&0.596&\textbf{0.626}&0.287&0.261&\textbf{0.406}\\
		BBC&0.569&0.569&\textbf{0.573}&\textbf{0.729}&\textbf{0.729}&0.728&\textbf{0.425}&\textbf{0.425}&0.421\\
		PRES&0.139&0.143&\textbf{0.151}&0.752&0.752&\textbf{0.765}&\textbf{0.396}&0.386&0.382\\
		MFF&0.255&0.254&\textbf{0.256}&0.46&0.455&\textbf{0.478}&\textbf{0.376}&0.372&0.354\\
		\noalign{\smallskip}\hline
	\end{tabular}
\end{table*}
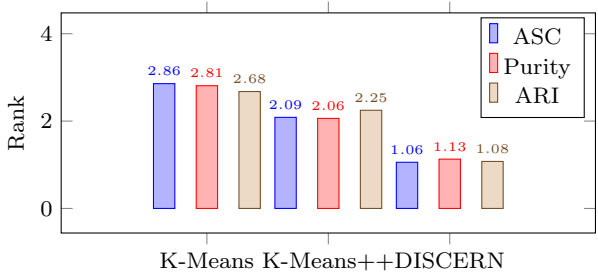
\begin{figure}[!ht]
	\centering
	\begin{tikzpicture}
	\begin{axis}[
	width=0.495\textwidth,
	height=4.5cm,
	ybar,
	enlargelimits=0.9,
	enlarge x limits=0.6,
	ylabel near ticks,
	ylabel=Rank,
	ybar=8pt,
	bar width=8pt,
	nodes near coords,
	nodes near coords style={font=\tiny},
	symbolic x coords={K-Means,K-Means++,DISCERN},
	xtick=data,
	]
	\addplot coordinates {(K-Means,2.8571)(K-Means++,2.0857)(DISCERN,1.0571)};
	\addplot coordinates {(K-Means,2.8095)(K-Means++,2.0619)(DISCERN,1.1286)};
	\addplot coordinates {(K-Means,2.6762)(K-Means++,2.2476)(DISCERN,1.0762)};
	
	\legend{ASC,Purity,ARI}
	\end{axis}
	\end{tikzpicture}
	\caption{Quade rankings of True K.}
	\label{fig:truekranks}
\end{figure}
\begin{table}[!ht]
	\centering
	\caption{Quade test Post-Hoc p-values using Li’s Method when using True K}
	\label{tab:truekpvalues}
	\begin{tabular}{cccc}
		\hline\noalign{\smallskip}
		Method&ASC&Purity&ARI\\
		\noalign{\smallskip}\hline\noalign{\smallskip}
		
		K-Means&\textbf{$2.8 \times 10^{-5}$}&\textbf{$9.2 \times 10^{-5}$}&\textbf{$1.94 \times 10^{-4}$}\\
		K-Means++&\textbf{$1.653 \times 10^{-2}$}&\textbf{$2.962 \times 10^{-2}$}&\textbf{$6.334 \times 10^{-3}$}\\
		
		\noalign{\smallskip}\hline
	\end{tabular}
\end{table}
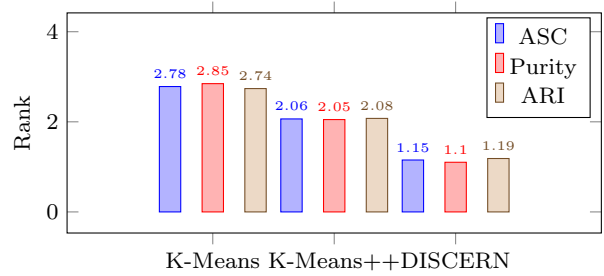
\begin{figure}[!ht]
	\centering
	\begin{tikzpicture}
	\begin{axis}[
	width=0.495\textwidth,
	height=4.5cm,
	ybar,
	enlargelimits=0.9,
	enlarge x limits=0.6,
	ylabel near ticks,
	ylabel=Rank,
	ybar=8pt,
	bar width=8pt,
	nodes near coords,
	nodes near coords style={font=\tiny},
	symbolic x coords={K-Means,K-Means++,DISCERN},
	xtick=data,
	]
	
	\addplot coordinates {(K-Means,2.7833)(K-Means++,2.0643)(DISCERN,1.1524)};
	\addplot coordinates {(K-Means,2.8476)(K-Means++,2.05)(DISCERN,1.1024)};
	\addplot coordinates {(K-Means,2.7381)(K-Means++,2.0762)(DISCERN,1.1857)};
	
	\legend{ASC,Purity,ARI}
	\end{axis}
	\end{tikzpicture}
	\caption{Quade rankings of DISCERN K.}
	\label{fig:discernkranks}
\end{figure}
\begin{table}[!ht]
	\centering
	\caption{Quade test Post-Hoc p-values using Li’s Method when using DISCERN K.}
	\label{tab:discernkpvalues}
	\begin{tabular}{cccc}
		\hline\noalign{\smallskip}
		Method&ASC&Purity&ARI\\
		\noalign{\smallskip}\hline\noalign{\smallskip}
		
		K-Means&\textbf{$1.49 \times 10^{-4}$}&\textbf{$4.9 \times 10^{-5}$}&\textbf{$3.09 \times 10^{-4}$}\\
		K-Means++&\textbf{$3.36 \times 10^{-2}$}&\textbf{$2.72 \times 10^{-2}$}&\textbf{$3.8 \times 10^{-2}$}\\

		\noalign{\smallskip}\hline
	\end{tabular}
\end{table}
In this subsection, we present the results of clustering on the datasets using 3 methods: K-Means, K-Means++ and DISCERN. We set the number of clusters to the number of ground truth classes for the \textbf{True K} experiment, and to the DISCERN-estimated number of clusters in the \textbf{DISCERN K} experiment.

We present the results of the clustering performance when the number of clusters is set to the number of ground truth classes in Table \ref{tab:truekcomparison}, and the results when the number of clusters is estimated by DISCERN in Table \ref{tab:discernkcomparison}. The resulting numbers are rounded to 3 decimal points. ASC is basically an indicator of how well an algorithm is able to cluster data in an unsupervised manner (regardless of the classes and the labels). Purity and ARI on the other hand, indicate how well the clustering algorithms have classified the datasets, when comparing cluster assignments to the ground-truth labels.
It is seen that DISCERN can usually reach better results than K-Means and K-Means++, especially when the representation is more suitable, i.e. face and image datasets.
In the cases that may contain noisy data, DISCERN may preform poorly compared to K-Means++, as it is more sensitive to noise. In such cases, DISCERN's deterministic nature will be its disadvantage as well, since K-Means++ has the potential to reach better results due to its stochastic nature.
We also conducted Quade statistical analysis on these results. The Quade statistic was calculated according to F-distribution with 2 and 26 degrees of freedom. Figure \ref{fig:truekranks} presents the rankings of the Quade test, and Table \ref{tab:truekpvalues} presents Li’s post-hoc p-values compared to DISCERN, since it was the top-ranked method in all three comparisons. The same analysis was conducted on the results from the \textbf{DISCERN K} experiment, and the results are presented in Figure \ref{fig:discernkranks} and Table \ref{tab:discernkpvalues}.
As it can be observed, the proposed method shows significant improvement in all metrics, as the p-values in both experiments are under the threshold of $5 \%$. Note that these results do not mean that the same standard is going to hold for all types of data, but rather sets of data similar to ones used in the experiments, all of which share one key feature: suitable feature representation.

\subsection{Clustering Stability}
\label{sec:stability}
DISCERN is deterministic, and therefore, unlike K-Means and K-Means++ which are stochastic, it does not require multiple runs of the algorithm in order to select the most preferable results. DISCERN could potentially perform better than the two while estimating the number of clusters, but may also suffer more complexity, which is further explained below. Furthermore, the stochastic nature of the other two is sometimes an advantage. DISCERN however has to remain deterministic for a suitable $K$ estimation.

\subsection{Complexity Analysis}
\label{sec:complexitycomparison}
In this section, we present the complexity order of DISCERN.
Based on Algorithm \ref{alg:discern}, DISCERN's complexity depends on whether the number of clusters is known or not. Assuming that the number of clusters($K$) is known, DISCERN's main loop runs for $K-2$ times which yields a total complexity of:
\begin{equation}
\label{eq:discerncomplexityK}
DISCERN(K) \in \bigO(nK^2 + dn^2)
\end{equation}
On the other hand, in the case where $K$ is unknown, DISCERN's complexity can be expressed as:
\begin{equation}
\label{eq:discerncomplexity}
DISCERN \in \bigO(n^3 + dn^2)
\end{equation}
As mentioned in Sect. \ref{sec:relatedwork}, K-Means has an order of $\bigO(ndKT)$ while K-Means++ initialization alone is going to add a complexity of $\bigO(ndK^2)$ to K-Means.
From Eq. (\ref{eq:discerncomplexityK}) we understand that DISCERN usually has a higher complexity than K-Means++, with a worst-case complexity of $\bigO(nK^2 + dn^2)$ when the number of clusters is known, but it can be more efficient than K-Means++ when the number of data points($n$) is much less than the dimension of the space($d$). An instance of that is term-document matrices in text mining.
We also provide the complexity of the elbow and silhouette methods below, and note that the following is based on the idea that no previous knowledge with respect to the data is available, therefore the limit for the number of clusters would be in the order of $n$. An instance is FEI, which contained $400$ images of about $200$ people ($= n/2$). In other words, these two methods run K-Means with a specific $K$ and increment that number each time and later evaluate which $K$ is more suitable. Therefore, even without the evaluation (score computation), $n-1$ runs of K-Means++($K$ starting at $2$ and ending at $n$) would yield a complexity which is expressed below:
\begin{flalign}
\label{eq:elbowcomplexitypre}
\begin{aligned}
&\text{K-ESTIMATION} \in \bigO\left( \sum\limits_{i=2}^{n}{ndi^2 + ndiT_i} \right)\\
&= \bigO\left(nd\sum\limits_{i=2}^{n}{i^2 + iT_i} \right)\\
&= \bigO\left(nd\frac{(n)(n+1)(2n+1)}{6}+nd\sum\limits_{i=2}^{n}{iT_i}\right)
\end{aligned}
\end{flalign}
Since $T_i$ is a constant each time, we exclude it from the complexity for simplicity.
\begin{flalign}
\label{eq:elbowcomplexity}
\begin{aligned}
&\text{K-ESTIMATION} \in\\
&\\
&\bigO\left(nd\frac{(n)(n+1)(2n+1)}{6}+nd\frac{(n)(n+1)}{2}\right)\\
&\\
&= \bigO\left(dn^4 + dn^3\right)
\end{aligned}
\end{flalign}
This yields that the worst-case complexity of the methods is $\bigO(dn^4)$, while DISCERN has a lower complexity of $\bigO(n^3 + dn^2)$.
This concludes that DISCERN has better performance than X-Means and is more efficient when compared with the silhouette and elbow methods.

\subsection{Summary and Discussion}
\begin{figure*}[!ht]
	\centering
	\includegraphics[width=0.8\textwidth]{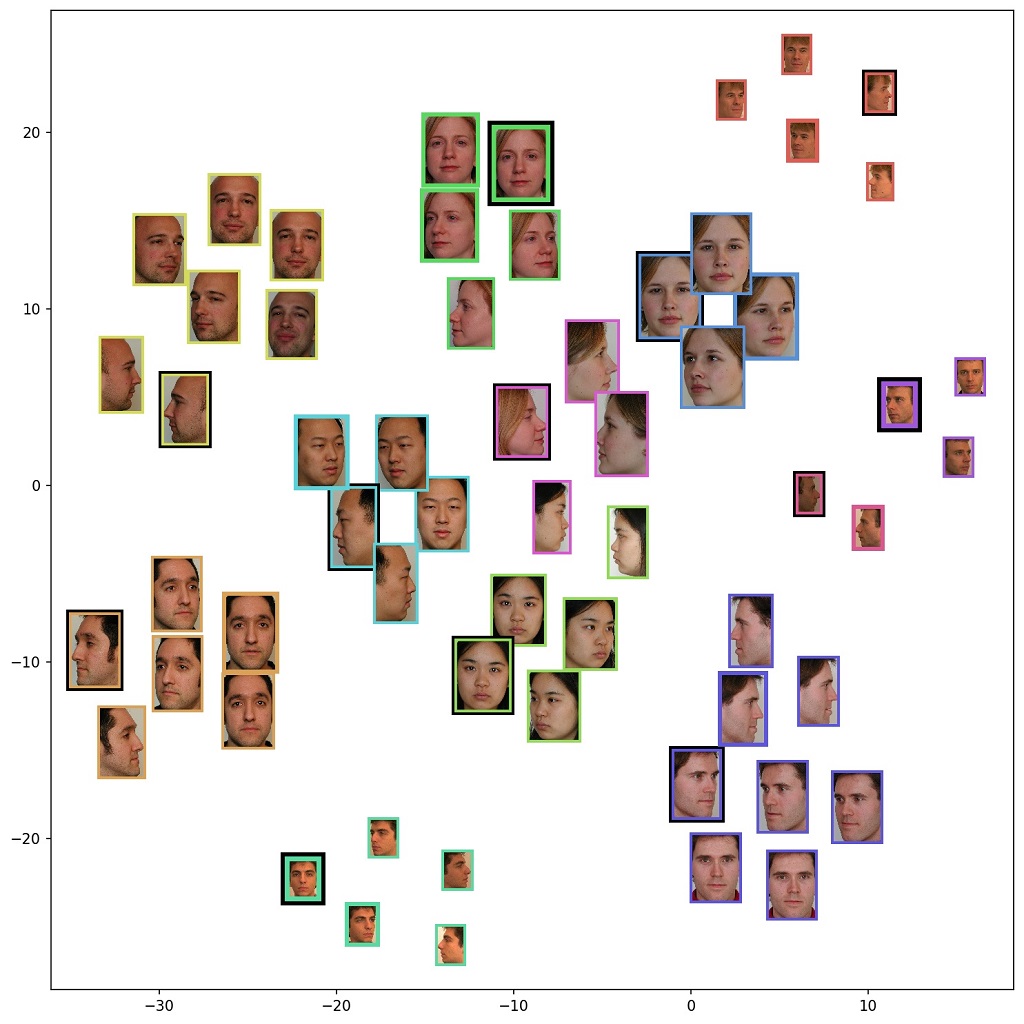}
	\caption{MIT clustered using DISCERN (t-SNE Visualization). Each cluster assigned is represented with a colored border, and the initial centroids have an extra black border.}
	\label{fig:discern_mit}
\end{figure*}

The proposed method values diversity, and picks the most diverse data points as centroids. This can be a bit problematic with the presence of noisy data. In the facial datasets specifically, we noticed that those with worse representations (rotated angles, dark lighting, and the like) were sometimes being clustered alone. A t-SNE visualization of the dataset MIT \cite{weyrauch2004component} being clustered by DISCERN is presented in Figure \ref{fig:discern_mit}. This dataset includes $10$ individuals each with a different number of images and from different angles.

While this can be a disadvantage in datasets with noisy data present, it serves also as the most obvious advantage in the rest.
To summarize, while K-Means and K-Means++ as partitional clustering measures are sensitive to noise, the proposed approach may be even more sensitive in specific cases. Datasets such as Yale \cite{belhumeur1997eigenfaces} on the other hand are perfectly clustered by DISCERN, with $100\%$ accuracy.
Another point worth noting is that estimating the number of clusters correctly requires features which to some degree hint at the correct number. For example, DISCERN estimated the number of clusters very closely to the number of classes in facial datasets, all of which went through a deep network which represents facial images better. As a result, DISCERN is highly efficient for well-constructed representations.

\section{Conclusion}
\label{sec:conclusion}
As discussed earlier, K-Means is an efficient clustering algorithm. Among many uses in data science, its most basic usage is data partitioning. However, it is very sensitive to initialization and K-Means++ has proven to be the most efficient initialization for K-Means by far.
Nevertheless, the problem of setting the number of clusters still exists in real-time applications. Person re-identification is a good example, in which a set of facial images or patterns are available, but the number of unique people is not.
Many methods such as X-Means, the elbow and silhouette methods have been used previously, but as seen in our experiments, they either fail to estimate a number close enough to the optimal number, or are inapplicable when there exists no knowledge of the minimum and maximum number of clusters. While it can be argued that these methods may be very useful tools for data scientists, they cannot possibly be built into a real-time service. These methods often require multiple runs of an initialization algorithm on top K-Means, which is highly inefficient. Moreover, these methods are using runs of an algorithm which is stochastic in order to obtain results, which may lead to instability in their overall results.
As a result, we introduced DISCERN which is an initialization algorithm that attempts to solve these issues.

DISCERN operates based on point-by-point similarity which is deterministic, therefore yields the same results. It chooses the most diverse data points as the initial centroids for K-Means.
This process can be thought of as a careful deterministic re-engineering of K-Means++, since the goal is essentially the same, while the selection process is made deterministic and adjusted to aide the estimation of the number of clusters. K-Means++ sets a selection probability for each data point which is relative to its diversity. DISCERN instead uses a different formulation for defining diversity (Eq. (\ref{eq:plcomputation})) which not only asserts diversity but also helps shape the function $R$ (Eq. (\ref{eq:membershipfunction})) which is later used to estimate the number of clusters. This entire process is done without runs of K-Means, which is part of the reason behind its lower complexity compared to methods such as the elbow and silhouette methods.
We compared DISCERN in both in estimating an optimal number of clusters, and as a clustering initialization method and found that it is ranked higher than K-Means++ in terms of suitable clustering, and ranked the best in terms of stability in results with an obvious standard deviation of zero. Moreover, it was the best performing method in terms of K-estimation and in terms of complexity order.
It should be added that our experiments in no way point toward the conclusion that DISCERN would perform better than K-Means++ in all cases. DISCERN can provide more suitable results in cases where noisy data doesn't exist in great capacities. This is the greatest weakness of DISCERN, as it is also the weakness of the original K-Means. Nevertheless, DISCERN would be even more affected by this issue than K-Means as it operates entirely based on diversity, which is arguably high among noisy data.
Density-based methods such as DBSCAN \cite{ester1996density} and OPTICS \cite{ankerst1999optics} can perform better in such cases where noise is present, but suffer from greater complexity compared to K-Means. Future research in this area can include measures that are diversity-based, but also take noise into consideration. An instance is employing neighborhood-based methods along with diversification. Spectral methods such as the method proposed by Little et al.  \cite{little2015multiscale} can also be helpful as they also change the embedding space.

Further improvements of this method may include a mini-batch version, or an online version of the method, as the computation of the similarity matrix can be very costly.
One future application of the proposed method is undoubtedly in deep learning. In our experiments, we used a pre-trained deep network \cite{schroff2015facenet, facenetkeras} trained with triplet loss in order to cluster facial data, and observed very good results even in the cases where the number of unique faces surpassed 100 (FEI). Following that, clustering methods can play an essential role in unsupervised and semi-supervised learning using deep learning, where deep learning finds a suitable feature representation, and the clustering is done afterwards, or even online clustering algorithms may be used to help deep networks learn representations in an unsupervised manner \cite{gansbeke2020learning, caron2020unsupervised}.

In summary, this method relies on a suitable feature representation which can be provided using matrix methods and deep learning, and when that representation is suitable, it serves as a completely parameter-less learning algorithm. This can serve as a step towards making machine learning algorithms more independent from human supervision.

\section*{Acknowledgments}
We would like to thank the anonymous reviewers for their valuable feedback and comments.
We also thank Dr. Farid Saberi Movahed for his useful comments and discussions.

\section*{Conflict of interest}
The authors declare that they have no conflict of interest.

\bibliographystyle{unsrtspbasic}      
\bibliography{references}   

\end{document}